\documentclass[fleqn,usenatbib]{mnras}

\usepackage{newtxtext,newtxmath}

\usepackage[T1]{fontenc}
\usepackage{ae,aecompl}

\usepackage{graphicx}	
\usepackage{amsmath}	
\usepackage{amssymb}	
\usepackage{float}
\usepackage{subcaption}

\usepackage{comment}

\title{Eliminating artefacts in Polarimetric Images using Deep Learning}

\author[D. Paranjpye et al.]{D. Paranjpye,$^{1}$\thanks{E-mail: pdhruv@umich.edu}
A. Mahabal,$^{2}$
A.N. Ramaprakash,$^{3}$
G. V. Panopoulou,$^{2}$
\newauthor
K. Cleary,$^{2}$
A.C.S. Readhead,$^{2}$
D. Blinov,$^{4}$
K. Tassis$^{4,5}$
\\
$^{1}$Department of Electrical and Computer Engineering, University of Michigan, 1301 Beal Ave, Ann Arbor, MI 48109,USA.\\
$^{2}$Division of Physics, Mathematics, and Astronomy, California Institute of Technology, Pasadena, CA 91125, USA.\\
$^{3}$ Inter-University Center for Astronomy and Astrophysics, Post Bag No. 4, Ganeshkhind, Pune - 411007, India.\\
$^{4}$ Department of Physics and Institute of Theoretical \& Computational Physics, University of Crete, Heraklion, Greece. \\
$^{5}$ Institute of Astrophysics, Foundation for Research and Technology - Hellas, Vassilika Vouton, GR-70013 Heraklion, Greece. \\ 
}

\date{Accepted XXX. Received YYY; in original form ZZZ}

\pubyear{2019}

\begin{document}
\label{firstpage}
\pagerange{\pageref{firstpage}--\pageref{lastpage}}
\maketitle
\begin{abstract}
Polarization measurements done using Imaging Polarimeters such as the Robotic Polarimeter are very sensitive to the presence of artefacts in images. Artefacts can range from internal reflections in a telescope to satellite trails that could contaminate an area of interest in the image. With the advent of wide-field polarimetry surveys, it is imperative to develop methods that automatically flag artefacts in images. In this paper, we implement a Convolutional Neural Network to identify the most dominant artefacts in the images. We find that our model can successfully classify sources with 98\% true positive and 97\% true negative rates. Such models, combined with transfer learning, will give us a running start in artefact elimination for near-future surveys like WALOP. 

\end{abstract}




\section{Introduction} \label{intro}

RoboPol  \citep{Ramaprakash2019} is a four-channel optical polarimeter installed on the 1.3m telescope at the Skinakas Observatory in Crete, Greece that is primarily used for polarimetry of point sources in the R band. Its successor the Wide Area Linear Optical Polarimeter (WALOP) is under development at the Inter-University Center for Astronomy and Astrophysics (IUCAA) in Pune, India. Images contain artefacts resulting from dust patterns, cosmic ray hits, satellite trails and pixel bleeding contaminating information from celestial objects. With the increasing number of images taken every night from such instruments, it is necessary to automate the analysis of data. However, with humans taken out of the loop it is possible for artefacts to get misidentified as a source and be used in the analysis. This would lead to erroneous results so the detection of such artefacts is imperative. 

Early work on detection of artefacts in astronomical images dates to the early 2000s when \cite{Storkey2004} used computer vision techniques such as the Hough Transform to detect linear artefacts like satellite trails, scratches, and diffraction spikes near bright stars. These methods were concerned with detection of linear features and highlighted some of the difficulties of using the Hough Transform  when dealing with light-density variations. 

Later on the focus shifted to object identification followed by extracting features for the objects and then classification using these features to separate out artefacts with methods like decision trees and random forests \citep{donalek2008new}. Recent years have seen the compilation of terrestrial datasets like ImageNet consisting of a million labelled images with a thousand categories  such as human faces, digits, vehicles, flowers, animals etc. \cite{deng2009imagenet} followed by the development of deep learning libraries and models using these datasets e.g. the VGG16 architecture \cite{simonyan2014very}.

With deep learning it is possible to skip the sometimes subjective step of feature extraction and go straight to classification after obtaining a labeled dataset  \citep[see e.g][]{Cabrera2017,duev2019RB,duev2019deepstreaks}. This is at the cost of explainability, but with proper validation and test datasets, the results are still reliable. Additional ways to improve the robustness, and faster convergence using techniques like Mask R-CNN and linear scaling combined with normalization are discussed in some recent papers such as \cite{he2017mask,gonzalez2018supervised,burke2019deblending}.

\begin{figure}
    \includegraphics[scale=0.33]{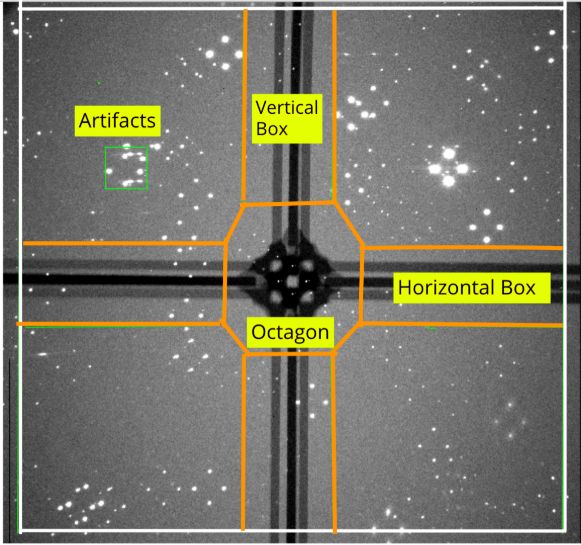}
    \caption{The region within the orange lines describes the restricted area from where we do not extract stars or artefacts. Similarly the regions outside the white lines are restricted areas. Note that the lines are exaggerated only for the purpose of representation.}
    \label{fig:mask}
\end{figure}

Our task here is to classify objects in RoboPol images into stars and artefacts. RoboPol images contain reflections of bright stars due to the interface between the 2 Wollaston Prisms used in the instrument and it is this dominant class of artefacts that we target here. The interface between the Wollaston Prisms is shown in the diagram of the optical instrument design described in \cite{Ramaprakash2019}, and an example of the reflection artefact in Figure \ref{fig:mask}. The green box in the upper left quadrant shows two horizontally extended artefacts separated vertically. A few stars in the vicinity also got included in the box.

In this paper, we propose to solve the problem of artefact detection for RoboPol images using an appropriately designed Convolutional Neural Network (CNN). 
In Section \ref{approach}, we introduce our approach to the problem of detecting artefacts in RoboPol images. We detail the implementation of our method including pre-processing steps, CNN architecture, and visualization of the output, and 
in Section \ref{discussion1} we discuss our findings and future possibilities.

\section{Approach}\label{approach}

 The RoboPol database consists of tens of thousands of images taken between 2013 and 2019. We first generate a data-set containing stars and artefacts and then develop a CNN to perform the classification. 
The following is an outline of our method:
\newline 1. Create training data for artefacts and stars from RoboPol images through manual labeling. This includes data for validation and testing. The manual labeling was done by visually inspecting about 100 images and recording the pixel coordinates of the artefacts.
\newline 2. Develop a CNN Architecture tuned through hyperparameter variation.
\newline 3. Train the model using training data obtained in step 1. 
\newline 4. Validate the model using validation and testing data.
\newline 5. Implement the model to find artefacts in an arbitrary RoboPol image. 

\subsection{Training Data for Artefacts and Stars}

Reflection artefacts and stars in RoboPol images have x and y extents from several pixels to a few tens of pixels. We chose a size of 64x64 pixels for our cutouts, with the artefacts and stars centered. Each star appears at four locations due to splitting of light from a single source within the instrument, with the locations lying at the vertices of a diamond. A detailed design implementation is available in \cite{Ramaprakash2019}.

For each image we generate a catalog of sources (including stars, reflection artefacts, and any other connected brightness peaks) using Sextractor\footnote{https:///sextractor.readthedocs.io}. We have roughly 10 artifacts and about 250 stars per image. The catalog comes with flags indicating various conditions such as saturation, proximity to another source, proximity to edge of image etc.\footnote{https://sextractor.readthedocs.io/en/latest/Flagging.html}.
About 70\% of the visually inspected artefacts had no error. This indicated that relying on just flags is not sufficient to separate artefacts.  To obtain training data of stars we make sure that from every image we extract stars of varying brightness and not just from a narrow brightness range. In each image, we chose this range to be 1 star per magnitude-bin for up to 5 magnitudes in each image. Likewise, our training data would contain an uniform distribution of magnitudes of brightness and ensure that we aren't biasing our neural network by providing training images from a limited magnitude range. We do not use all sources so that the sets of stars and artefacts can stay roughly equal, and hence balanced for the classification process. 

Unlike typical astronomy images, RoboPol images contain a mask (Figure \ref{fig:mask}). More details about the mask can be found in \cite{Ramaprakash2019}. 
The code repository and documentation are available at  https://github.com/delta-papa/Robopol-artifacts

\subsection{CNN Architecture}

\begin{figure*}
    \centering
    \includegraphics[width=0.99\textwidth]{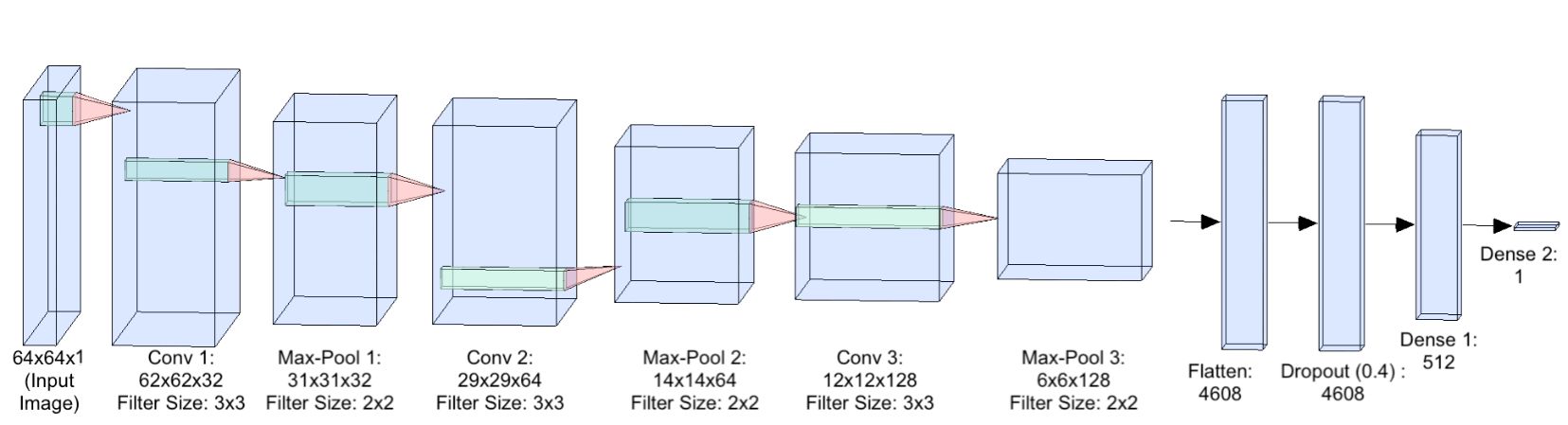}
    \caption{The CNN model consists of 3 convolutional layers, 3 max-pooling layers, 2 fully connected layers and a dropout layer for regularization. In case of the convolutional and max-pooling layers, the size of the layer and the filter size used are also mentioned. For the dense layers, the total number of nodes are shown. The input is a 64x64 pixel PNG image.}
    \label{fig:architecture}
\end{figure*}

We follow the now standard image classification model developed by the Visual Graphics Group (VGG) at Oxford, UK \citep{simonyan2014very}. Our implementation uses 3 convolution layers, 3 max-pooling layers and 2 fully connected layers (see Figure \ref{fig:architecture}). The hidden layers are activated using a ReLU (Rectified Linear Unit) activation. Finally we use a sigmoid activation at the output layer. The first, second and third convolution layer consist of 32, 64 and 128 filters respectively each with a kernel size of 3x3 and stride length of 1. The max-pooling layers use a kernel size of 2x2 pixels. At the end of the 3rd max-pooling layer we use a dropout layer with a probability of dropping a node as 0.4 for regularization ensuring no single parameter of the neural network has a very high coefficient \citep{srivastava2014dropout}. The total number of trainable parameters in our configuraton are 2,452,993 and we use an Adam optimizer to perform back-propagation \citep{kingma2014adam}. The loss function used is a binary cross-entropy loss. 

\subsection{Data Augmentation and Training}

While going through the images we saw that most of the artefacts are due to internal reflections and had a horizontal streak-like shape. For proper training of the neural network we need to generate a large data-set of training images. Therefore, to augment the number of images for training, we rotated the cutouts by 180 degrees so that the horizontal nature of the artefacts is preserved. 

\begin{figure*}
    
    \centering
    \begin{subfigure}{0.45\textwidth}
        \centering
        \includegraphics[width=0.9\textwidth]{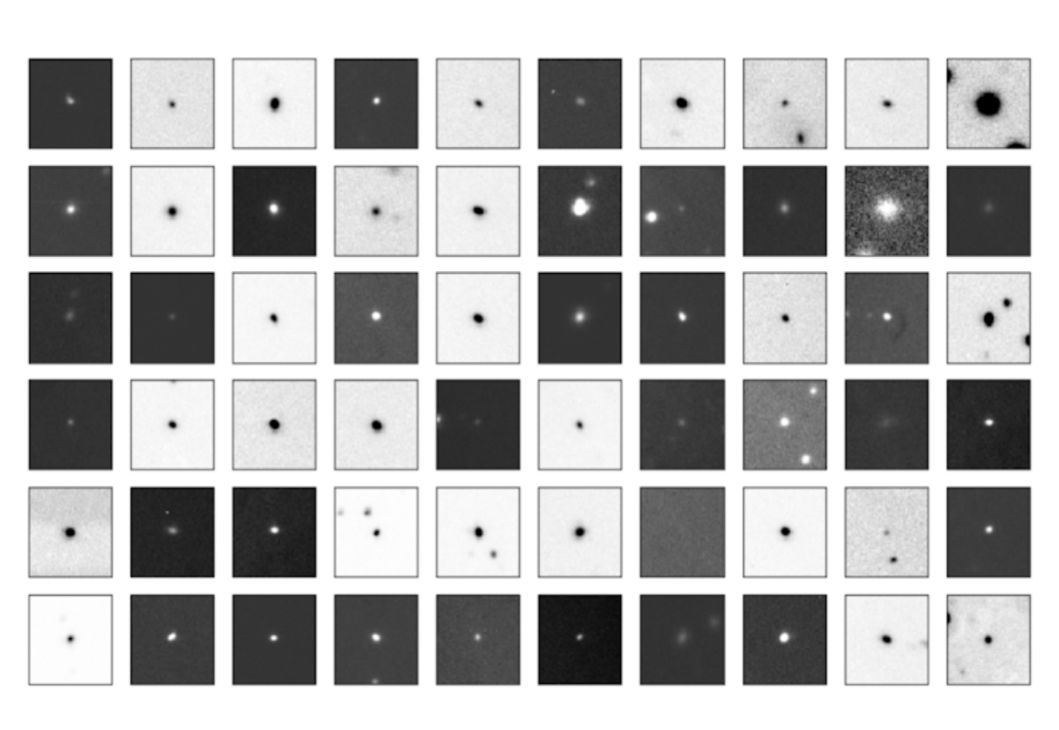} 
        \caption{Cutouts of Stars used as training images.}
    \end{subfigure}\hfill
    \begin{subfigure}{0.45\textwidth}
        \centering
        \includegraphics[width=0.9\textwidth]{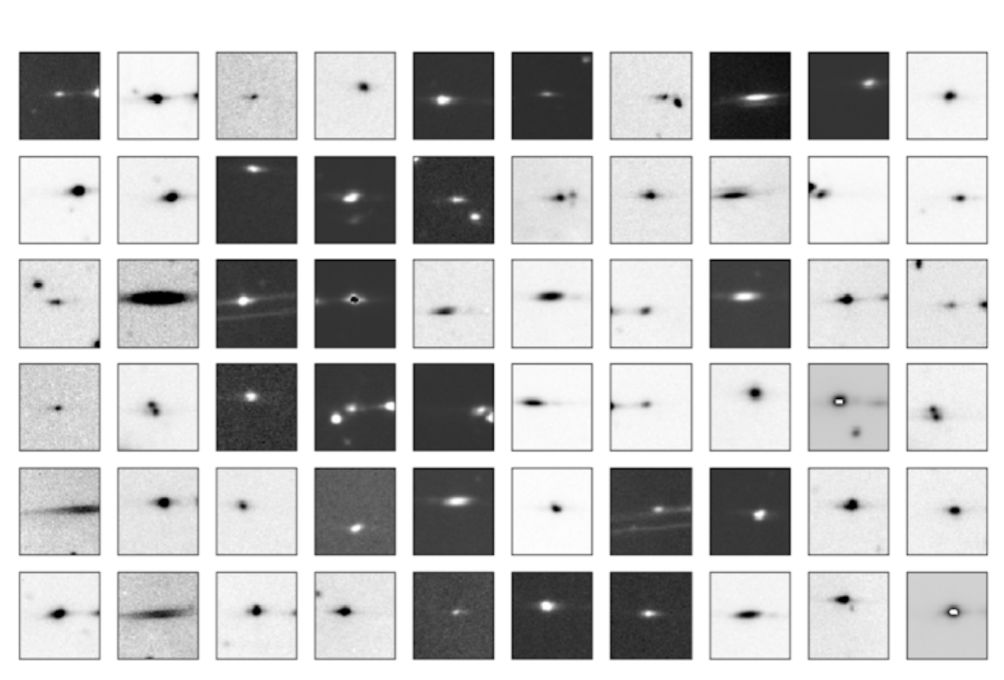} 
        \caption{Cutouts of Artefacts used as training images.}
    \end{subfigure}
    \caption{Training images for stars (a) and artefacts (b).}
\end{figure*}

To split the data into training and validation we used a 80-20 ratio and shuffled the data-set randomly while splitting to avoid bias. For training, we had a total of  836 images of stars and 925 images of artefacts. The training data was augmented using the ImageDataGenerator class in the high-level Keras\footnote{www.keras.io} API of Python. We performed horizontal and vertical flipping, width and height shifts and shearing. 
The shifts were applied to account for possible inaccuracies in centering of the samples.
The validation and training images were both normalized to [0,1] by dividing by 255, the maximum value of an 8 bit image. 

\begin{figure}
    \includegraphics[width=0.45\textwidth]{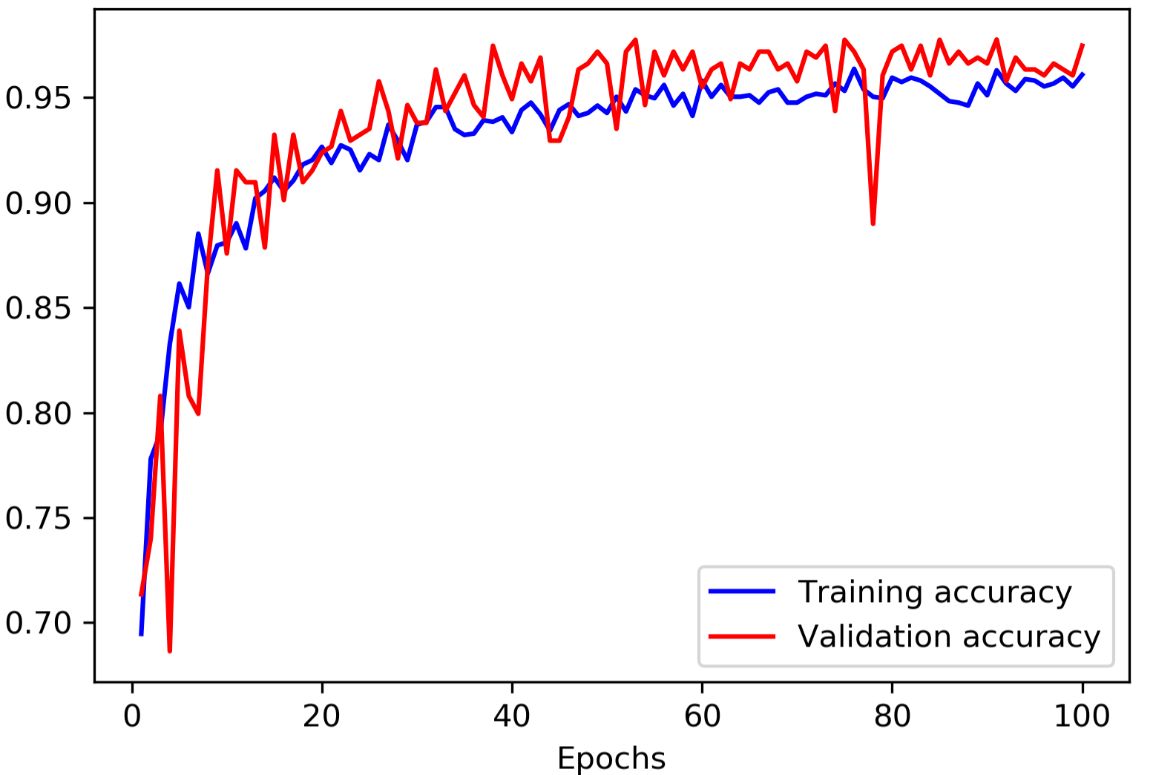}
    \caption{Training and Validation Accuracy for the final model.}
    \label{fig:accuracy}
\end{figure}

The hardware used for performing training was a 2.3 GHz Intel Core i5 processor and the total training time was 40 minutes. 

\subsection{Training Performance}
The total number of images chosen for training the model was 1408 (80\% of the 1761 images), and the remaining 353 images were reserved for validation. Training data was used to update the parameters of the model while validation data was used to only evaluate the model's performance after every update. The batch size used was 4. A total of 100 epochs were used in the training and the steps per epoch was set to 1408/4 i.e. 352. We used a learning rate of 0.001. The training accuracy reached about 95\% while the validation accuracy was close to 96\% at the end of 100 epochs as seen in Figure \ref{fig:accuracy}.

\begin{figure}
    
    \includegraphics[scale=0.17]{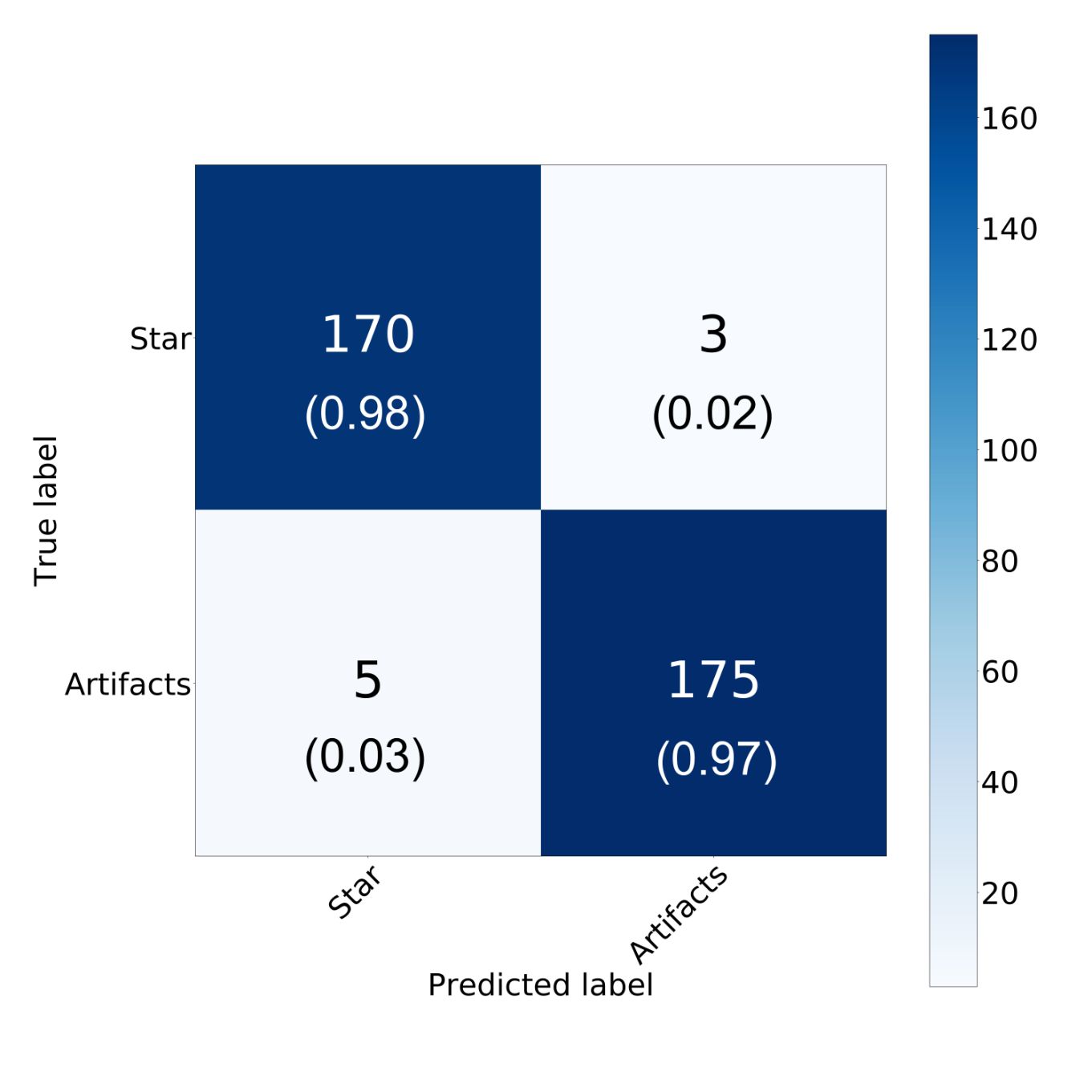}
    \caption{Confusion Matrix for the Binary Classification performed using our model. The normalized numbers are given in brackets. Out of 353 images, 170 were True Positives, 175 were True Negatives, 5 False Positives and 3 False Negatives. The False Positive Rate is 3\% and False Negative Rate is 2\%.}
    \label{fig:confusionmatrix}
\end{figure}

Besides making small changes to the hyperparameters above, we also implemented a network with 2 and 4 convolution layers to see whether there was any advantage in using shallower (2 layers) or deeper (4 layers) CNNs. The training accuracy and validation accuracy reached about 90\% in the shallower network while it reached about 96\% in the deeper network. Although training and validation accuracy may be good indicators of the proper working of a CNN in a binary classification problem there are other important parameters we need to consider when the costs of misclassification are high. For example, we are interested in knowing the false positive rate (sources wrongly classified as stars), the false negative rate (sources falsely classified as artefacts), as also the precision (fraction of sources correctly classified) and recall (fraction of stars correctly classified).  These numbers are summarized through two metrics viz. F1 score and Matthew's Correlation coefficient, and indicate whether our model is working as expected or not.

We need our system to have a high precision and recall score and the F1 score summarizes the 2 scores by taking their harmonic mean. 
The Matthew's Correlation Coefficient (MCC) is akin to a correlation coefficient measure between the predicted labels and true labels. A value of +1 indicates perfect positive correlation between the 2 quantities. Equations 
\ref{prec}, \ref{recall}, \ref{f1} and \ref{mcc} give the formulae for Precision, Recall, F1 score and MCC respectively. Note that TP, FP, TN, FN stand for True Positives, False Positives, True Negatives and False Negatives respectively.

\begin{equation}\label{prec}
    Precision = \frac{TP}{TP + FP}
\end{equation}

\begin{equation}\label{recall}
    Recall = \frac{TP}{TP + FN}
\end{equation}

\begin{equation}\label{f1}
    F1\ score = \frac{2 \cdot precision\cdot recall}{precision+ recall}
\end{equation}

\begin{equation}\label{mcc}
    MCC = \frac{TP*TN - FP*FN}{\sqrt{(TP+FP)(TP+FN)(TN+FP)(TN+FN)}}
\end{equation}

\begin{figure}
    
    \includegraphics[scale=0.2]{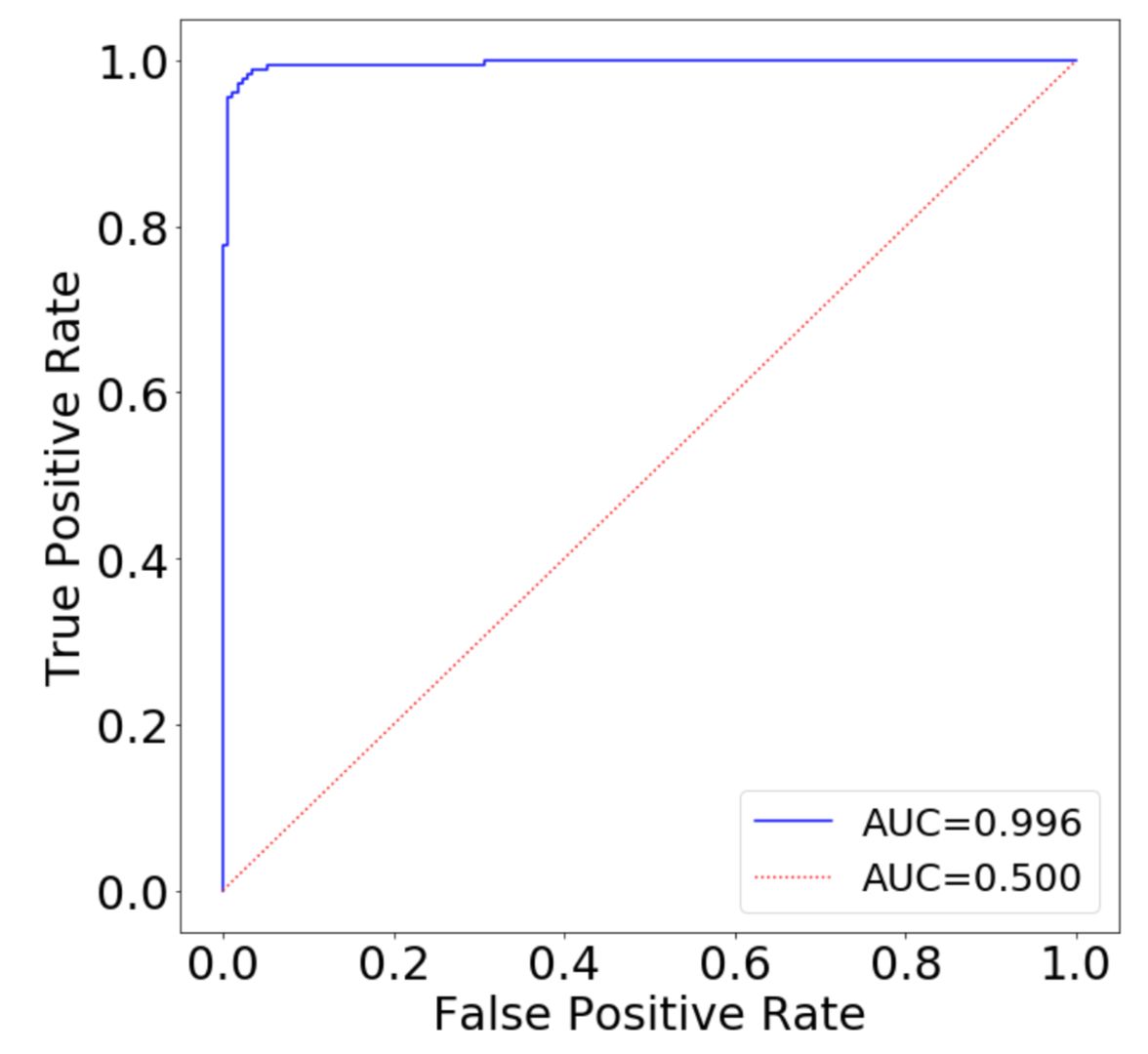}
    \caption{ROC Curve for our final model.}
    \label{fig:roc}
\end{figure}

\begin{figure}
    \includegraphics[scale=0.2]{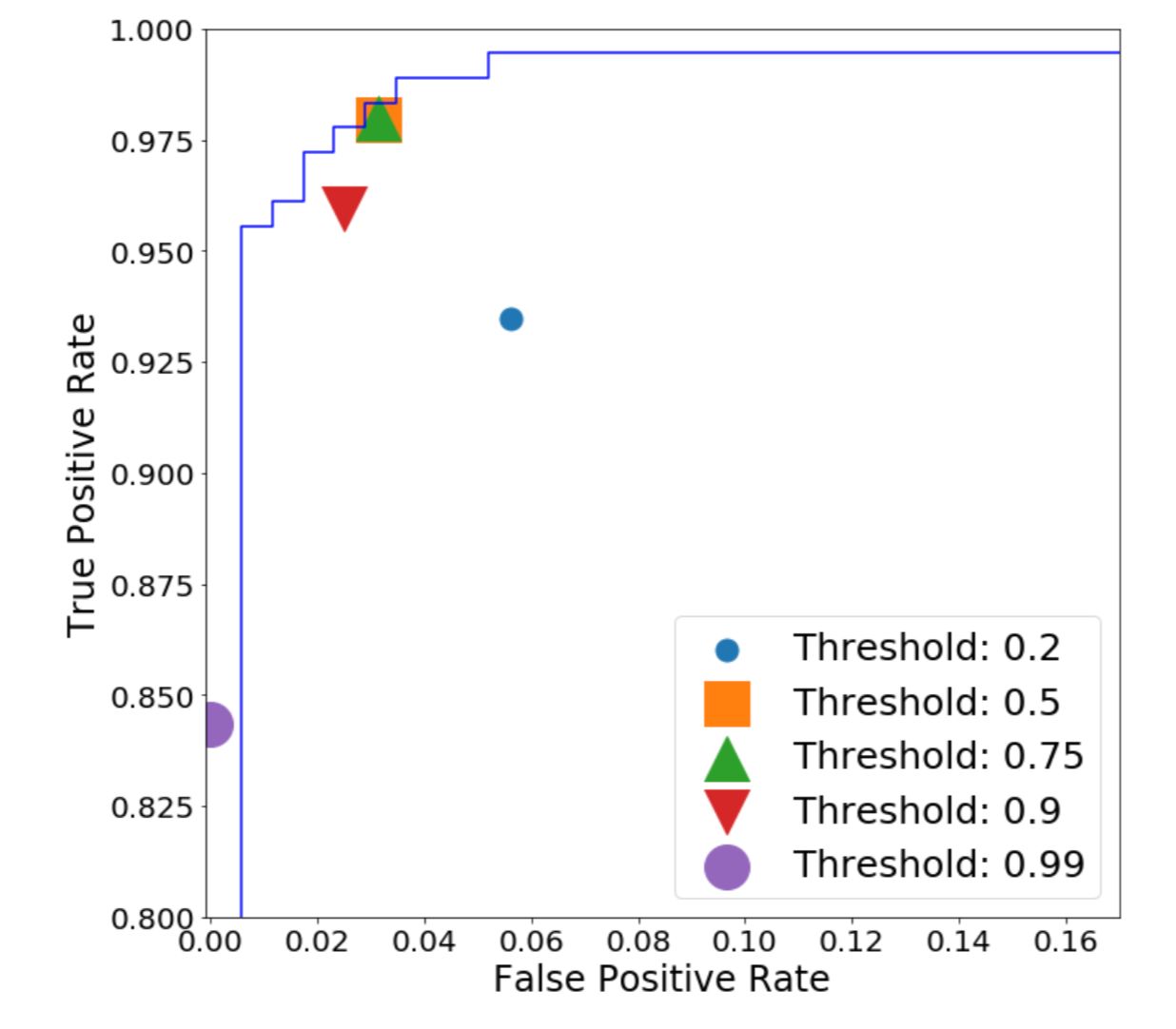}
    \caption{Zoom in of ROC (Figure \ref{fig:roc}) to highlight the non-ideal area.}
    \label{fig:roczoom}
\end{figure}

\begin{figure}

    \includegraphics[scale=0.16]{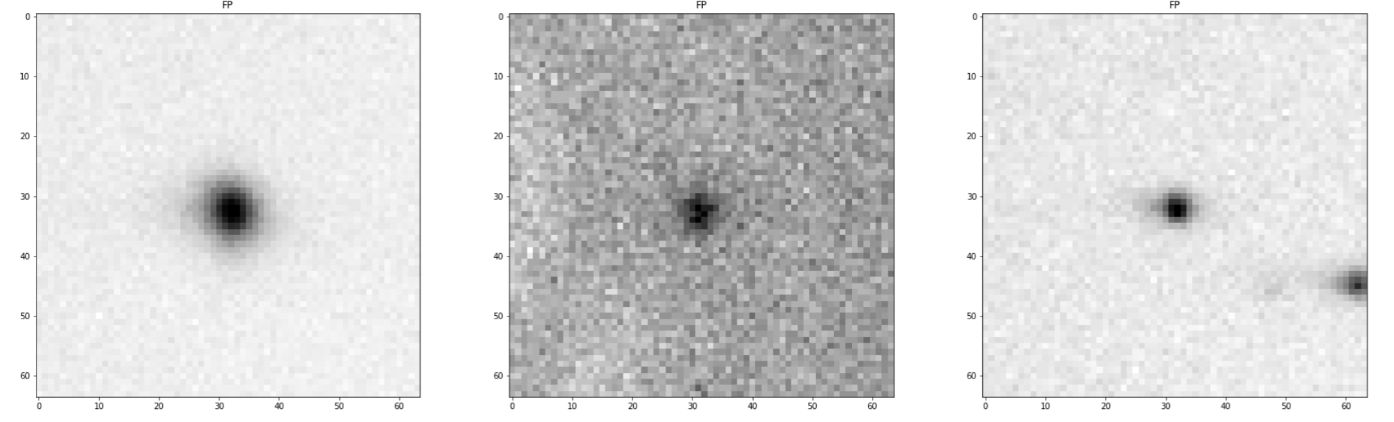}
    \caption{Some of the False Negative classifications where stars were classified as artefacts.}
    \label{fig:fn}
\end{figure}

\begin{figure}
    \includegraphics[scale=0.16]{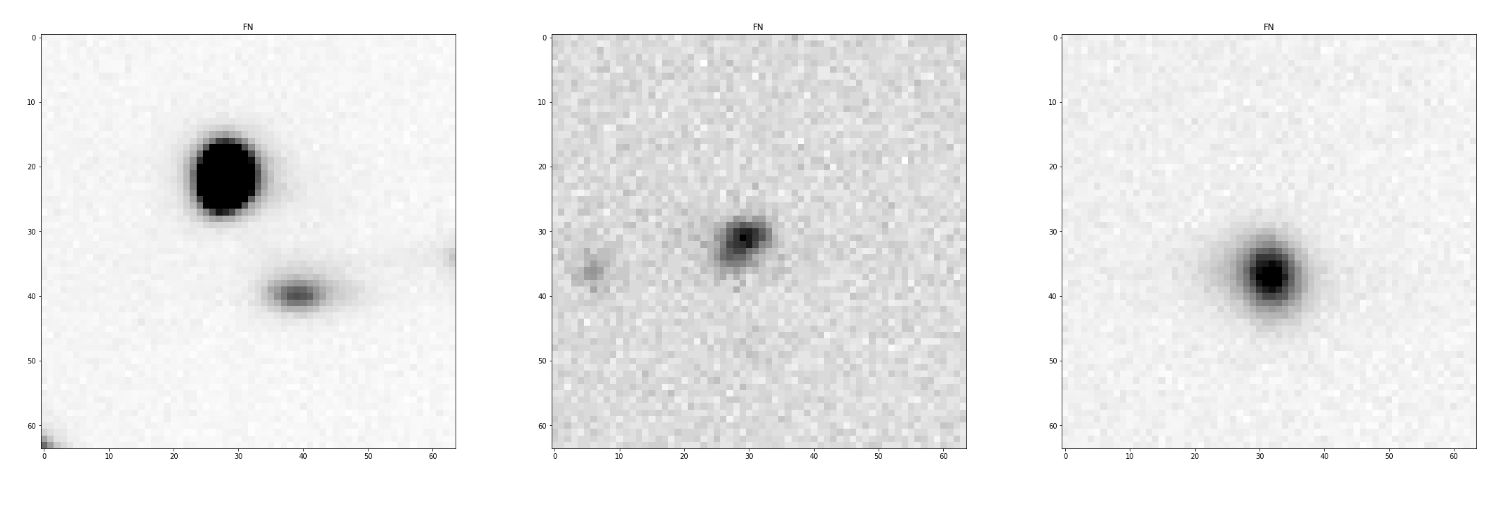}
    \caption{Some of the False Positive Classifications where artefacts were classified as stars. We see that the artefacts here didn't have a prominent streak-like shape. Also, the first artefact has a star near the center and the artefact is slightly away from the center.}
    \label{fig:fp}
\end{figure}

Figures \ref{fig:confusionmatrix} and \ref{fig:roc} show the confusion matrix and ROC curve respectively. The Confusion Matrix tells us the number of true positives, true negatives, false positives and false negatives. The Receiver Operating Characteristics (ROC) curve is a measure of the trade-off between the true positive rate and false positive rate for different values of the threshold used in the classifier. The threshold is a value between 0 and 1. If the probability of the source being a star is greater than the value of the threshold we classify the source as a star else as an artefact. Our threshold is set to 0.5. Ideally, we want the false positive rate to be 0 and true positive rate to be 1. Ideally we expect our ROC curve to have an area (under the curve) to be equal to 1. The Area Under the Curve (AUC) for our model is 0.996 while that for a random classifier is 0.5 as shown by the red dotted line. A zoom-in of Figure \ref{fig:roc} is shown in Figure \ref{fig:roczoom} which shows the values True Positive Rates and False Positive Rates at different thresholds. It shows that our threshold of 0.5 coincides with 0.75 indicating that our classifier can confidently achieve the same True Positive Rate and False Positive Rate at a higher threshold. Table \ref{table:comparison} compares the performance of our model with a shallower and deeper neural network. Based on this comparison, we chose the model with 3 convolution layers as the model with 4 layers actually shows a reduction in F1 score and MCC with a deeper and hence computationally more expensive network, possibly due to over-fitting. Figures \ref{fig:fn} and \ref{fig:fp} show some of the false negatives and false positives. 

\begin{table}
\centering
\caption{Performance comparison of CNN models with different layers. Here FPR is False Positive Rate, FNR is False Negative Rate, MCC is Matthew's Correlation Coefficient.}
\label{table:comparison}
\begin{tabular}{lccc}
 \hline
Parameter &2 Layers &3 Layers& 4 Layers \\
 \hline
 Training Accuracy & 0.90 & 0.95 & 0.90 \\
 Validation Accuracy & 0.91 & 0.96 & 0.95 \\
 FPR & 0.07 & 0.03 & 0.02 \\
 FNR & 0.13 & 0.02 & 0.05 \\
 Precision & 0.88 & 0.98 & 0.95 \\
 Recall & 0.93 & 0.97 & 0.98 \\
 F1 Score & 0.90 & 0.98 & 0.96 \\
 MCC & 0.80 & 0.95 & 0.93\\
 \hline
\end{tabular}
\end{table}

\subsection{Testing and Implementation}

\begin{figure}
    \includegraphics[scale=0.3]{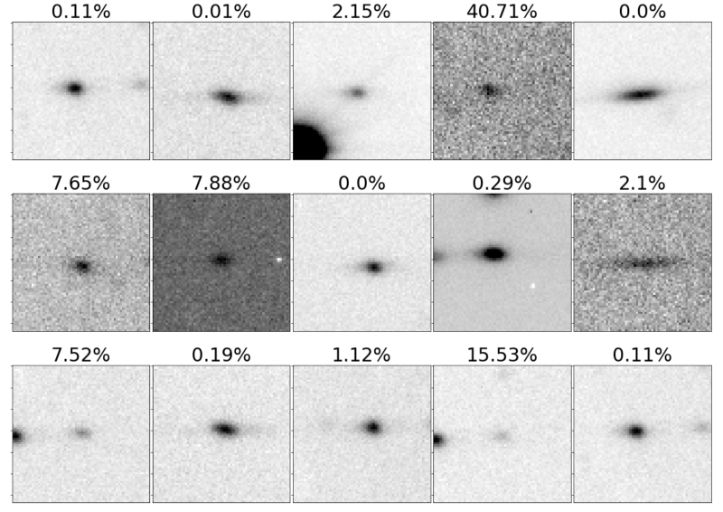}
    \caption{Result of testing the classifier on images of artefacts. All artefacts have been classified correctly with high classification probabilities. The classification probability of a source being a star is shown on the top of each cutout.}
    \label{fig:arti_result}
\end{figure}

\begin{figure}
    \includegraphics[scale=0.30]{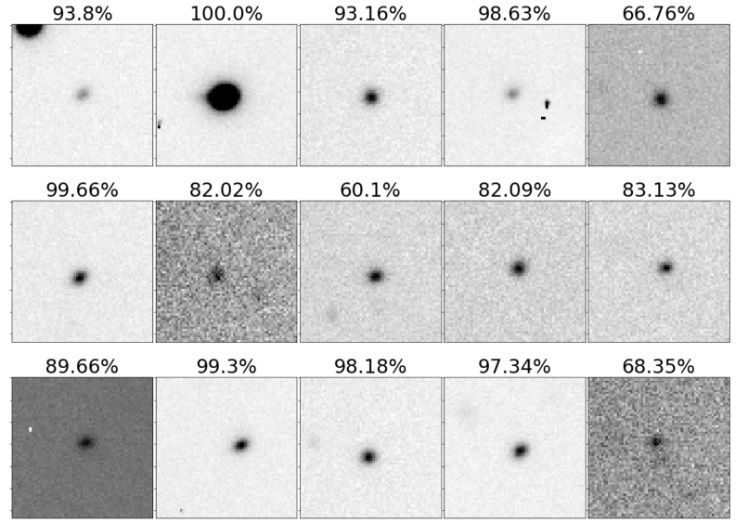}
    \caption{Result of testing the classifier on images of stars. The classification probability of a source being a star is shown on the top of each cutout.}
    \label{fig:star_result}
\end{figure}

\begin{figure}
    \centering
    \includegraphics[scale=0.25]{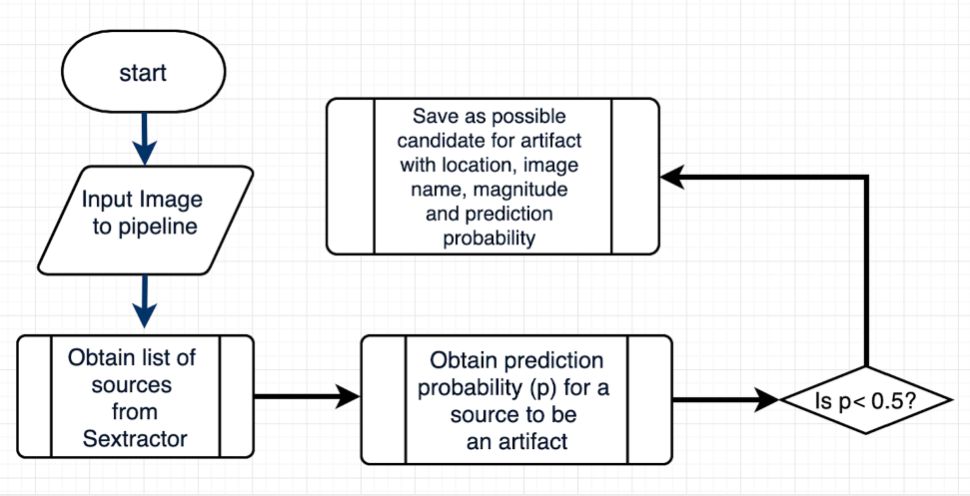}
    \caption{A flowchart showing the decision logic of the pipeline.}
    \label{fig:flowchart}
\end{figure}

\begin{figure}
     \includegraphics[scale=0.18]{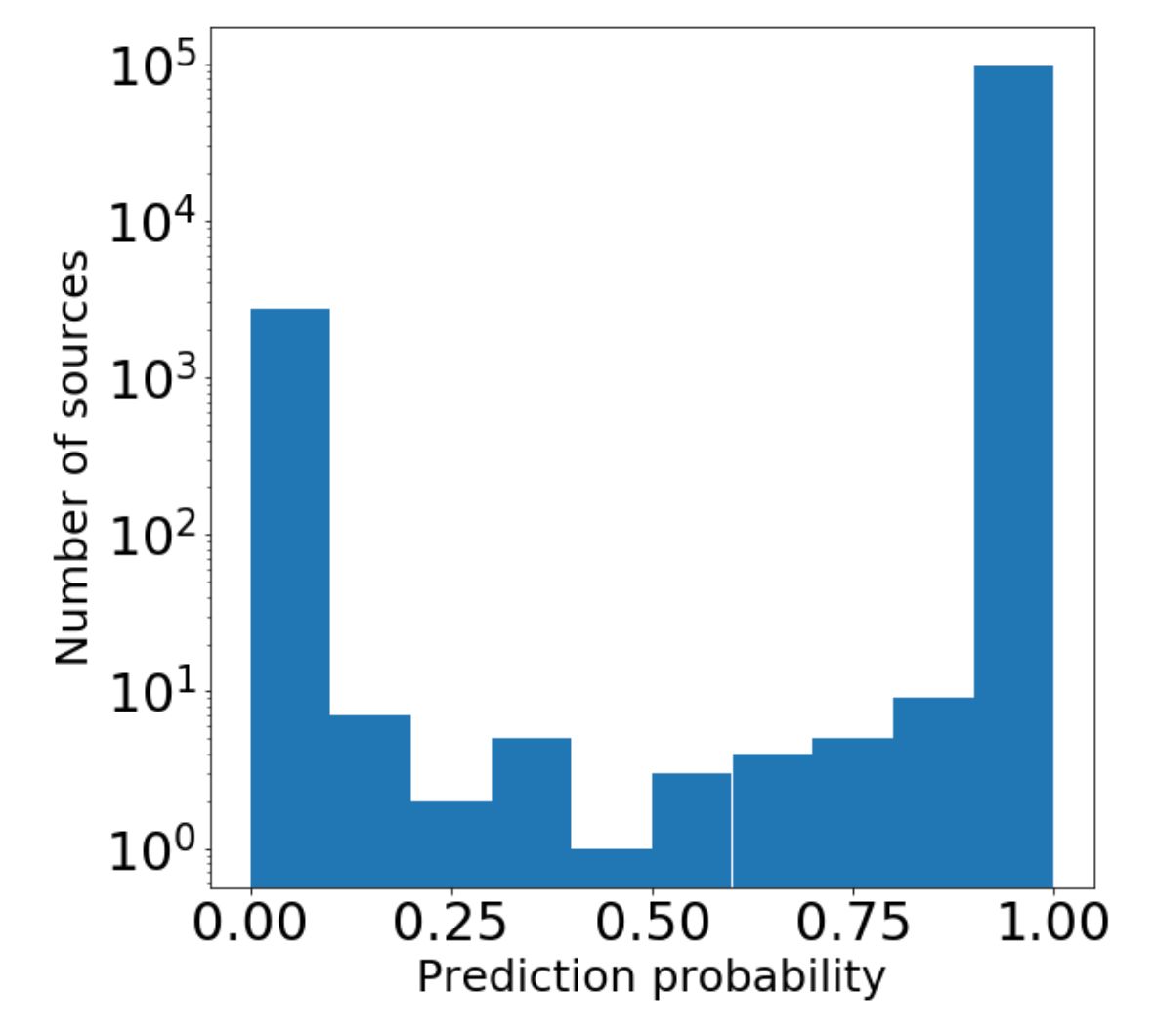}
     \caption{Histogram of prediction probabilities for ~91,000 sources.}
     \label{fig:hist1channel}
 \end{figure}

 We used our model to test 100 randomly chosen images - distinct from the training set - from the 40,000 images of RoboPol taken during the years 2013 and 2014. Our goal was to classify all the sources in each of the 100 images into stars and artefacts and analyze the results. For each image, we obtained  a list of sources, their positions, instrumental magnitudes and extraction flag error by using Sextractor. A histogram of the predicted probability of the sources being stars is shown in Figure \ref{fig:hist1channel}. Out of ~91,000 sources, we have ~88,000 sources classified as stars (90 to 100\% probability of being a star) and 2500 sources classified as artefacts (0 to 10\% probability of being a star). The inset plot shows that in the remaining prediction probability range there are fewer than 10 objects in each bin of size 10\%. This means that $\sim0.05\%$ of the sources had probabilities in the range between 10 to 90\%. 
 
 Figures \ref{fig:arti_result} and \ref{fig:star_result} show the results of classification on test images. Each image contains a single source with known label. At the top of each image is the probability of the source being a star. Sources in Figure \ref{fig:arti_result} are artefacts while those in Figure \ref{fig:star_result} are stars. In both categories, our classification rates are almost always above 90\%. 

The implementation pipeline takes an input image and the corresponding Sextractor file as its arguments and produces a list of locations of the detected artefacts along with their location marked in the original image with an associated probability. A decision logic diagram is shown in Figure \ref{fig:flowchart}. 

\subsection{Visualization with Saliency Maps}\label{viz}
\begin{figure}
    \includegraphics[scale=0.22]{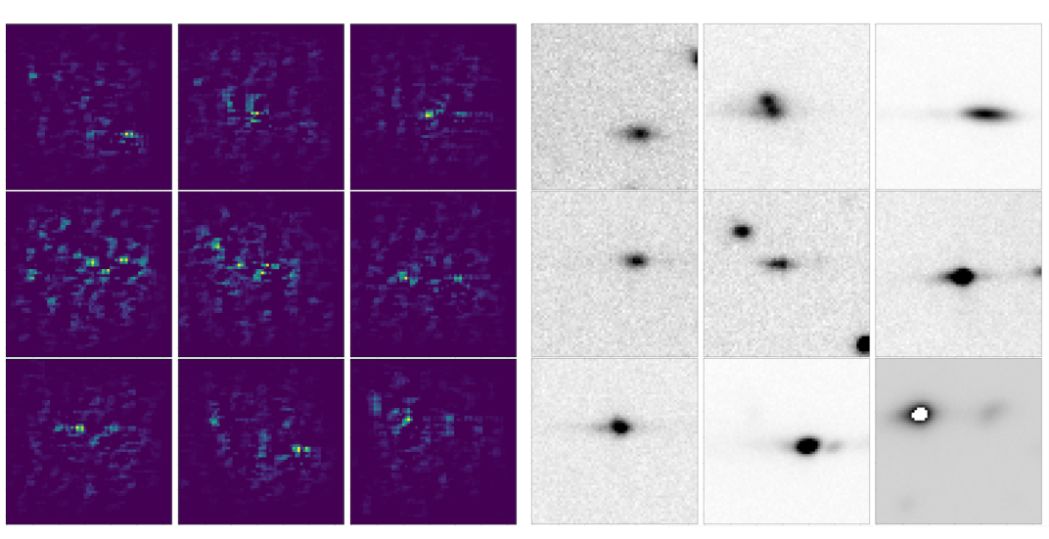}
    \caption{Saliency Map visualization (left) for 9 different input images (right) of artefacts. The Saliency map shows the positive gradients of the image with respect to the artefact class score and thus the locations of the artefacts in the image. These maps show that indeed the classifier is activated by the artefacts themselves and not their background. Figure plotted using Keras Visualization Toolkit \citep{raghakotkerasvis}.}
    \label{fig:saliency}
\end{figure}

A saliency map helps us find the locations of the pixels in  input images which need to be changed the least to activate the output filter. This means we find the gradient of the input image with respect to the output score. To visualize a saliency map, positive gradients are chosen that would give us the locations of the pixels activating the output filter. In other words this gives us the location of the object of the relevant class in the input image. A saliency map thus gives us the salient features of the class-specific input image that maximize the class score. A detailed mathematical treatment can be found in \cite{journals/corr/SimonyanVZ13}. Figure \ref{fig:saliency} shows the saliency maps for 9 different input images. Observe that for the image in row 3, column 3, only the artefact is visualized in the saliency and not the star at the top right corner of the image. Sextractor's ellipticity measure alone is not sufficient to separate the artefacts.

\section{Discussion}\label{discussion1}
\begin{figure}
    \includegraphics[scale=0.36]{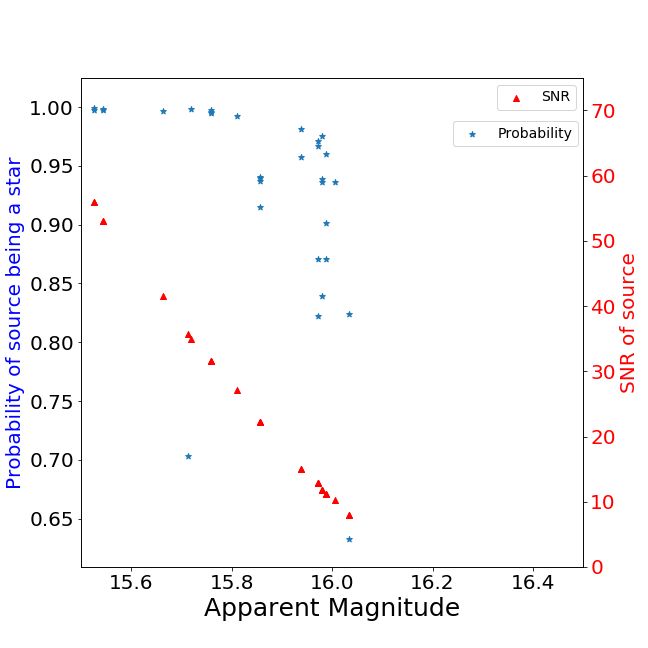}
    \caption{Scatter plot of prediction probability and signal to noise ratio (SNR) of visually inspected stars from a single image versus the apparent R1 magnitude.  The 42 stars span a magnitude range from 12 to 16.}
    \label{fig:magscatter}
\end{figure}

We used Convolutional Neural Networks to solve the problem of detecting artefacts in polarimetric images. Although the use of CNNs in astronomical image classification is not new, this is the first time that they have been used for detecting artefacts in polarimetric images. The efficiency of the method shows its suitability for use in upcoming polarimetry surveys such as the Polar Areas Stellar Imaging in Polarization High Accuracy Experiment \citep{pasiphae}, which will use the novel Wide Area Linear Optical Polarimeter (WALOP). Our implementation suggests that this method can be reliably used for detecting other kinds of artefacts as well given enough training data. The RoboPol instrument operates down to 16th magnitude in the R1 band. Figure \ref{fig:magscatter} shows that our Deep Learning model can classify stars down to 15.9 magnitude with a prediction probability better than 0.9. We have also plotted the signal to noise ratio (SNR) of the stars on a separate axis. We see that our model works up to SNRs of 15. Thus, our implementation works well with objects within the magnitude range RoboPol observes.  

In this paper, we do not use the spatial correlation for stars appearing as a diamond pattern in RoboPol images. That is because the diamond structure in each image of RoboPol is specific to the RoboPol polarimeter design and wouldn't be present in a single image of future polarimeters such as WALOP.

In the RoboPol data-set, we had majority of artefacts due to scattering of light from off-axis stars at the interface of the Wollaston prisms \citep{Ramaprakash2019}. Our method demonstrates that a binary classifier trained on images of stars and artefacts can successfully differentiate between them.  Our training data does not contain enough examples for artefacts such as satellite trails or bleeding pixels and as a result deep learning them is non-trivial without aggressive data augmentation. There already exist methods to remove such artifacts. Out-of-Distribution detection networks \citep{huang2019out} can also be used to detect such infrequent outliers. The final pipeline can incorporate such methods to deliver artefact-free products. 

\section*{Acknowledgements}
The work has been funded by the National Science Foundation under the NSF grant (161547). AM acknowledges support from the NSF (1640818, AST-1815034) and IUSSTF (JC-001/2017). KT acknowledges support  from  the  European Research Council  under  the  European  Union’s Horizon  2020 research  and  innovation  program,  under  grant  agreement  No771282.




\bibliographystyle{mnras}
\bibliography{mybib} 





\bsp	
\label{lastpage}
\end{document}